\documentclass{article}
\usepackage{amssymb}
\usepackage{booktabs}
\usepackage{multirow}
\usepackage{xcolor}
\usepackage{colortbl}
\usepackage{amsmath}
\usepackage{graphicx}
\usepackage{wrapfig}
\definecolor{mygray}{gray}{.85}
\definecolor{verylightgray}{RGB}{240,240,240} 
\usepackage{listings}
\usepackage{color}
\usepackage{algorithm}
\usepackage{algorithmic}
\usepackage[framemethod=TikZ]{mdframed}  
\newmdenv[  
  backgroundcolor=verylightgray,  
  hidealllines=true,  
  innerleftmargin=8pt,  
  innerrightmargin=8pt,  
  innertopmargin=2pt,  
  innerbottommargin=4pt  
]{graybox}  

\usepackage[preprint]{corl_2025} 

\title{GC-VLN: Instruction as Graph Constraints for Training-free Vision-and-Language Navigation}

\author{
  Hang Yin$^{123}$\thanks{~Equal contribution. $^\dag$ Project lead. $^\ddagger$ Corresponding author.}, Haoyu Wei$^{123*}$, Xiuwei Xu$^{123\dagger}$, Wenxuan Guo$^{123}$, Jie Zhou$^{123}$, Jiwen Lu$^{123\ddagger}$ \\
  $^1$Department of Automation, Tsinghua University \\
  $^2$Beijing Key Laboratory of Embodied Intelligence Systems \\
  $^3$Beijing National Research Center for Information Science and Technology \\
}

\begin{document}
\maketitle

\vspace{-5mm}
\begin{abstract}
In this paper, we propose a training-free framework for vision-and-language navigation (VLN).
Existing zero-shot VLN methods are mainly designed for discrete environments or involve unsupervised training in continuous simulator environments, which makes it challenging to generalize and deploy them in real-world scenarios.
To achieve a training-free framework in continuous environments, our framework formulates navigation guidance as graph constraint optimization by decomposing instructions into explicit spatial constraints.
The constraint-driven paradigm decodes spatial semantics through constraint solving, enabling zero-shot adaptation to unseen environments.
Specifically, we construct a spatial constraint library covering all types of spatial relationship mentioned in VLN instructions.
The human instruction is decomposed into a directed acyclic graph, with waypoint nodes, object nodes and edges, which are used as queries to retrieve the library to build the graph constraints.
The graph constraint optimization is solved by the constraint solver to determine the positions of waypoints, obtaining the robot's navigation path and final goal.
To handle cases of no solution or multiple solutions, we construct a navigation tree and the backtracking mechanism.
Extensive experiments on standard benchmarks demonstrate significant improvements in success rate and navigation efficiency compared to state-of-the-art zero-shot VLN methods.
We further conduct real-world experiments to show that our framework can effectively generalize to new environments and instruction sets, paving the way for a more robust and autonomous navigation framework. \href{https://bagh2178.github.io/GC-VLN/}{Project Page.}
\end{abstract}    
\section{Introduction}
\label{sec:intro}

Vision-and-language navigation (VLN)~\cite{mattersim} is an essential foundational capability for various embodied tasks, which requires the robot to move in a novel environment following complex linguistic instructions.
The VLN instructions typically describe the path sequence, including the direction and distance of movement along the path, as well as scene information near the path.
This necessitates the robot's ability of linguistic comprehension, environment perception, and spatial reasoning.
However, limited by the availability of annotated data, early data-driven VLN methods exhibit poor generalization ability in unseen scenarios.
Moreover, most existing data-driven VLN methods~\cite{an2024etpnav,chen2022weakly,zhang2024navid} are trained in simulator environment, which leads to a large sim-to-real gap.
Therefore, building data-independent zero-shot navigation frameworks is necessary.

Zero-shot VLN methods~\cite{zhou2023navgpt,chen2024mapgpt} have been proposed to overcome the limitation of annotated data and narrow the sim-to-real gap.
DiscussNav~\cite{long2024discuss} leverages multi-expert discussions to integrate instruction understanding, environmental perception, and decision verification.
MapGPT~\cite{chen2024mapgpt} builds a map-guided GPT-based agent that leverages an online linguistic-formed map for adaptive path planning.
However, constrained by immature simulators, these zero-shot VLN methods can only operate in discrete environments~\cite{mattersim}, which only allows the robot to move between a set of discrete nodes within the environment.
Methods in discrete simulator environment are impractical for deploying in real-world scenarios, introducing a significant sim-to-real gap. 
Vision-and-language navigation in continuous environments (VLN-CE)~\cite{krantz_vlnce_2020} is introduced to better simulate real-world environments, where the robot can move freely and stand at any position within the environment.
In VLN-CE, the robot predicts low-level actions, including turning and moving forward, which is closer to real-world environments.
The VLN-CE methods can be easily deployed in novel real-world scenarios without the adaptation.
Nevertheless, most existing zero-shot VLN-CE methods still rely on unsupervised training within simulators.
A$^2$Nav~\cite{chen2023a2navactionawarezeroshotrobot} constructs five subtasks for VLN-CE and trains the five subtask modules in a self-supervised manner within the simulator.
The self-supervised training on simulator data means that these methods still exhibit a sim-to-real gap.
Therefore, a training-free framework for VLN-CE is highly demanded.

In this paper, we propose \textbf{G}raph-\textbf{C}onstraints for \textbf{V}ision-and-\textbf{L}anguage \textbf{N}avigation (GC-VLN), a training-free framework for VLN-CE.
Different from previous works which is deployed in discrete environments or self-supervised trained on simulator data, GC-VLN adopts a completely training-free approach to perceive the scene and predict the path.
Specifically, we construct a constraint library to cover all types of spatial relationship constraints found in the instructions.
Then the linguistic instruction is decomposed into a directed acyclic graph, which is used to query the library to construct a graph constraint.
The navigation is formulated as a graph constraint optimization problem, where the coordinates of the waypoint nodes are progressively solved by the constraint solver.
Additionally, we build a navigation tree to address the problem of uncertainty in the number of solutions during graph constraint solving.
We conducted extensive experiments in the simulator benchmarks R2R-CE and RxR-CE, achieving state-of-the-art performance across the board. Real-world experiments further demonstrate the strong generalization ability of our method.

\begin{figure*}
  \centering
    \includegraphics[width=\linewidth]{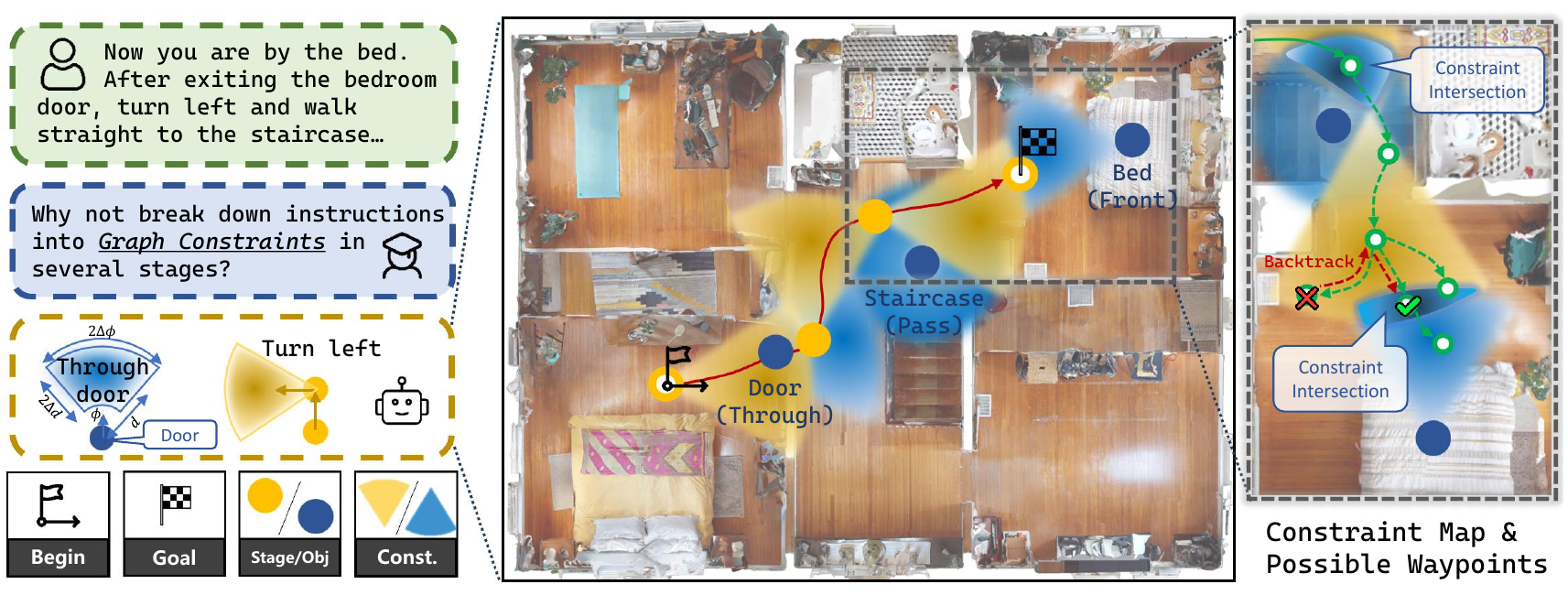}
  \caption{
  GC-VLN models the instructions as a graph constraint optimization problem and solve the graph constraints based on the robot's observations to obtain the robot's path, which enables training-free VLN. We illustrate how graph constraints guide the navigation path and how to re-plan the path when exploration fails.}
  \vspace{-3mm}
  \label{fig:teasor}
\end{figure*}
\section{Related Work}

\subsection{Vision-and-Language Navigation}

In real-world navigation scenarios, humans typically have prior knowledge of the path to the goal.
Therefore, robots do not need to autonomously explore the goal.
Using human prior knowledge, the robot follows human instructions to reach the goal, a task called Vision-and-Language Navigation (VLN)~\cite{mattersim,hao2020prevalent,anderson2021sim,chen2021topological}.
Early VLN methods~\cite{liu2023bird,long2024discuss} use discrete simulator environments to evaluate VLN approaches, where the environment is divided into multiple discrete drivable waypoints.
The robot can only stand on the waypoints and selects one of the adjacent waypoints as a short-term goal at each step.
Discrete environments allow the method to focus on path selection.
However, due to the large sim-to-real gap, discrete VLN methods are difficult to deploy in real-world scenarios.

Towards the aim of closely approximating real-world scenarios, Vision-and-Language Navigation in Continuous Environments (VLN-CE)~\cite{krantz_vlnce_2020} is proposed, which allows the robot to move freely within the scene by predicting low-level actions.
Methods designed~\cite{krantz2022sim, wang2023dreamwalker, wang2023gridmm, Wang_lookahead, an2024etpnav} for VLN-CE can be directly deployed in real-world scenarios without modification.
To balance the simplification of methods in discrete VLN with the approximation of real environments in VLN-CE, the waypoint predictor~\cite{Hong_2022_CVPR} is proposed to predict discrete waypoints online within a continuous environment, and becomes the mainstream technical approach for VLN-CE task.

\subsection{Training-free Navigation}

Conventional navigation methods are task-training, involving modules such as LSTM~\cite{hochreiter1997long} and transformer~\cite{vaswani2017attention}.
The limit of annotated data results in these methods having restricted generalization abilities for more diverse goals or human instructions.
Moreover, task-training methods also exhibit sim-to-real gap, which limits their performance in real-world scenarios.

Zero-shot navigation~\cite{yu2023l3mvn, yin2024sgnav} methods have been proposed to address the generalization problem.
In the field of goal-oriented navigation, zero-shot methods for object-goal navigation~\cite{zhou2023esc, cai2023bridgingzeroshotobjectnavigation, wu2024voronav, yu2023l3mvn, kuang2024openfmnav}, image-goal navigation~\cite{guo2025iglnav, krantz2023navigating, yin2025unigoal, wei2024ovexp}, and text-goal navigation~\cite{sun2024prioritizedsemanticlearningzeroshot} have already reached a relatively mature stage.
For VLN in discrete environments, zero-shot methods~\cite{li2024tinathinkinteractionaction} have also seen significant development.
However, for the VLN-CE task, zero-shot methods remain unsatisfactory, which is the goal of our method.
\section{Approach}

\begin{figure*}
  \centering
    \includegraphics[width=\linewidth]{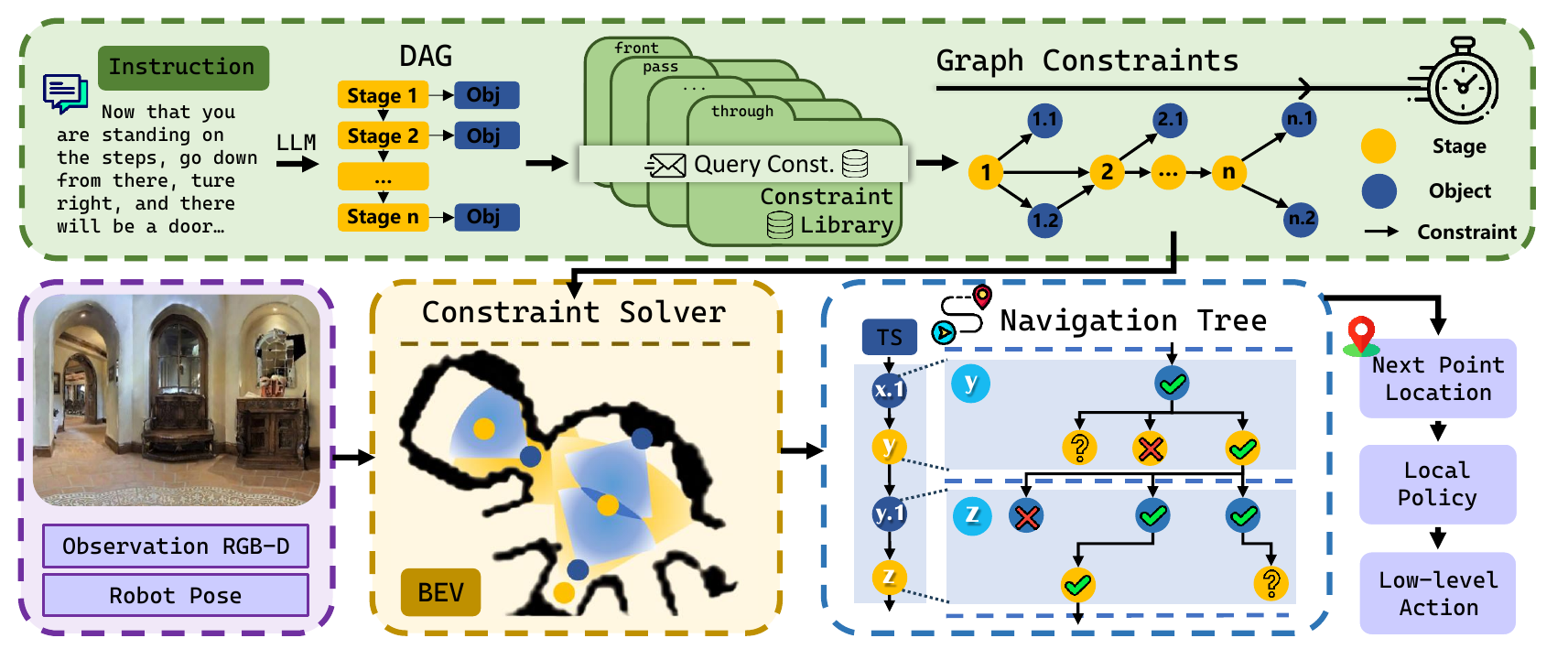}
  \caption{Framework of GC-VLN. We construct a constraint library, containing all the spatial relationship mentioned by navigation instruction. The instruction is decomposed into a directed acyclic graph (DAG) and used to query the library to get the graph constraints. The constraint solver determines the path by solving the graph constraint optimization. According to the topological sort (TS) of graph constraint, we build the navigation tree, where the number of leaf nodes equals the number of graph constraint solutions. In graph constraint and TS, $t.i$ means the $i$-th object node in stage $t$.}
  \label{fig:pipeline}
\end{figure*}

We first present the task definition and the overall pipeline of our approach, followed by an explanation of how the graph constraint $\mathcal{K}$ is constructed.
Finally, we elaborate on how we sequentially solve the coordinates of nodes in $\mathcal{K}$ and utilize $\mathcal{K}$ to guide the navigation direction.

\subsection{Overview}

In VLN, a robot is initialized in an unknown environment.
The robot is required to follow the linguistic instruction $\mathcal{I}$ provided by a human, so that it can move within the environment and reach the final destination.
As shown in Figure \ref{fig:teasor}, the instruction is typically a piece of text that describes how to navigate from the starting point to the destination point, including navigation directions, objects encountered along the navigational path, and spatial relationships between the navigation path and objects.
If the agent reaches within $r$ meters of the navigation endpoint in no more than $t$ steps, the navigation is successful.

\textbf{Pipeline.}
As shown in Figure \ref{fig:pipeline}, the pipeline of GC-VLN contains two main modules, the graph constraint construction module and constrained optimization module.
First, the original instruction is decomposed into a multi-stage directed acyclic graph $\mathcal{G}$, which contains all the information required for navigation.
We construct a constraint library that encompasses all types of spatial relationships in VLN instruction.
The graph $\mathcal{G}$ is used to query this library to obtain the constraint types between nodes, thereby the graph constraint $\mathcal{K}$ is constructed.
The determination of the node coordinates in $\mathcal{K}$ is formulated as a constrained optimization problem based on their constraints, and the navigation tree handles the uncertain number of coordinate solutions from the constraint solver, to backtrack when no solution meets the constraint conditions.

\subsection{Graph Constraint Construction}

\begin{wrapfigure}{r}{0.5\linewidth}
  \centering
    \includegraphics[width=\linewidth]{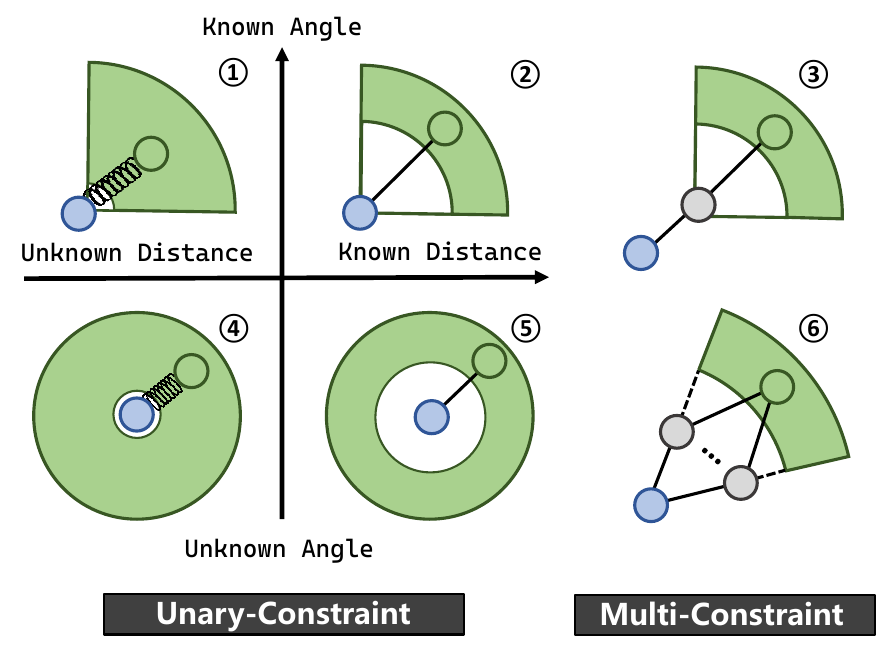}
  \caption{Diagram of the Constraint Library containing six types of constraint. For constraint $c\left(v \mid u\right)$, $u$ and $v$ are colored blue and green, respectively. The green region is the possible region for $v$.}
  \label{fig:constraint}
\end{wrapfigure}

To handle the long sequence characteristics and complex spatial relationships in the instruction $\mathcal{I}$, we convert $\mathcal{I}$ into a structured representation, namely the graph constraint, which is required to meet three criteria:
1. It must not lose any information from $\mathcal{I}$.
2. It must explicitly contain all the objects mentioned in $\mathcal{I}$.
3. It must provide explicit navigation directions, as well as spatial relationships between objects and the navigation path.

\textbf{Instruction Decomposition.}
The LLM is prompted to decompose the instruction $\mathcal{I}$ into multiple navigation stages, where each stage involves exactly one displacement (no requirement for the number of rotations).
Each stage has two attributes: the navigation direction and the objects that appear in that stage.
The navigation direction is categorized as one of the following: \texttt{["front", "right", "left", "back", "unknown"]}.
Every object mentioned in $\mathcal{I}$ belongs to and belongs only to one navigation stage.
Each object has one attribute: its spatial relationship with the navigation path.
These nodes and edges constitute a directed acyclic graph $\mathcal{G}$.

Graph $\mathcal{G}=\text{LLM}(\mathcal{I})$, where $\mathcal{G}=(\mathcal{V}, \mathcal{E})$, and $\mathcal{V},\mathcal{E}$ represent the nodes and the directed edges, respectively.
$\mathcal{V}$ can be categorized into waypoints $\mathcal{V}^{w}$ and object nodes $\mathcal{V}^{o}$, where $\mathcal{V}^{w}=\left\{v_{1}^{w}, v_{2}^{w}, \ldots, v_{n}^{w}\right\}, \mathcal{V}^{o}=\bigcup_{i=1}^{n}\left\{v_{i 1}^{o}, v_{i 2}^{o}, \ldots, v_{i k_{i}}^{o}\right\}$. Edges can be categorized into $\mathcal{E}^{w}$ and $\mathcal{E}^{o}$ based on whether they are connected to an object node, where $\mathcal{E}^{w}=\left\{\left(v_{i}^{w}, v_{i+1}^{w}\right) \mid 1 \leq i \leq n-1\right\},\mathcal{E}^{o}=\bigcup_{i=1}^{n}\left(\left\{\left(v_{i}^{w}, v_{i j}^{o}\right) \mid 1 \leq j \leq k_{i}\right\} \cup\left\{\left(v_{i j}^{o}, v_{i}^{w}\right) \mid 1 \leq j \leq m_{i}\right\}\right)$.
$v_{i}^{w}$ and $v_{ij}^{o}$ represent the waypoint nodes and the $j$-th object nodes in stage $i$, and $e=(u,v)$ represents the edge pointing from node $u$ to node $v$.
For details of instruction decomposition, please refer to the supplementary materials.

\textbf{Constraint Library.}
The spatial relationships between nodes mentioned in $\mathcal{I}$ are considered spatial constraints $c\left(v \mid u\right)$ of their coordinates $u,v$.
The type of constraint is distinguished based on the description of the spatial relationship topology in $\mathcal{I}$.
As shown in Figure \ref{fig:constraint}, we construct a constraint library $\mathcal{L}$ containing six types of constraint, which covers all types of spatial relationship topologies involved in VLN instructions.

Assuming $u$ is known, the second type of constraint as an example, the constraint $c\left (v \mid u\right )$ can be expressed as:
\begin{align}
    c\left(v \mid u\right)&=(c^{a}\left(v \mid u\right),c^{d}\left(v \mid u\right)), \quad \text{sum}(c)=c^{a}+c^{d}, \quad \text{min}(c)=\text{min}(c^{a},c^{d}) \\
    c^{a}\left(v \mid u\right)&=\cos (\Delta \phi)\|v-u\|-[\|v-u\|-(v-u) \cdot(\cos \phi, \sin \phi)]\\
    c^{d}\left(v \mid u\right)&=\Delta d^{2}-(\|v-u\|-d)^{2}
\end{align}
where $\phi,\Delta \phi$ and $d,\Delta d$ are the baseline and tolerance of angle and distance of $e$, and $c^{a},c^{d}$ are the sub-constraints of angle and distance.
The complete expressions of the sub-constraints of the other five types of constraint are provided in the supplementary materials.

\textbf{Graph Constraint.}
We query the constraint library $\mathcal{L}$ using $ e \in \mathcal{E} $ to obtain the type of constraint $c=\mathcal{L}(e)$ corresponding to $e$.
All nodes and constraints together form the graph constraint $\mathcal{K}$:
\begin{equation}
    \mathcal{K}=(\mathcal{V},\mathcal{C}),\quad \mathcal{C}=\{\mathcal{L}(e) \mid e \in \mathcal{E}\}
\end{equation}
The direction of $c$ represents the direction of causality and inferring, meaning that the coordinates of $v$ can only be inferred when the coordinate of its parent node $u$ is known.
At the beginning of the navigation, the coordinate of $v_{1}^{w}$ is $(0,0)$ and that of other nodes in $\mathcal{V}$ are all uncertain.
At the end of the navigation, all coordinates of $v_{i}^{w}\in\mathcal{V}^{w}$ are determined and $v_{n}^{w}$ is the endpoint of this episode.

\subsection{Constrained Optimization for Node Coordinates}

The graph constraint $\mathcal{K}$ is used to guide the robot's direction of motion. Therefore, the coordinates of the nodes $v \in \mathcal{V}$ are determined according to the navigation order. We propose a graph-constrained optimization framework to determine the coordinates and a navigation tree to handle the uncertainty in the number of solutions.

\textbf{Constraint Solver.}
Graph constraint $\mathcal{K}$ is utilized to solve the coordinates of waypoints $v_{i}^{w} \in \mathcal{V}^{w}$.
We determine the coordinates of object nodes $v_{ij}^{o} \in \mathcal{V}^{o}$ by perceiving RGB-D observations with a pre-trained vision model.
Solving for the coordinate of a node $v$ requires knowledge of the coordinates of all its parent nodes in $\mathcal{K}$.
In the initial state of navigation, only the starting point of stage 1, which serves as the root node $v_{1}^{w}$ of $\mathcal{K}$, has known coordinates.
$v_{1}^{w}$ acts as the original reference for solving the coordinates of all subsequent nodes.
We perform a topological sort (TS) on $\mathcal{K}$ to determine the order in which the node coordinates are solved during navigation.
The order of the topological sort ensures that for any constraint $c\left (v \mid u \right )$, $v$ always follows after $u$.
Therefore, when determining the node coordinates in this order, the coordinates of the current node's parent nodes are guaranteed to be already solved.
The details of the topological sort can be found in the supplementary materials.

For the waypoint $v_{i}^{w}$, multiple constraints $\mathcal{C}_{v_{i}^{w}}=\{c\left(v_{i}^{w} \mid v_{i-1}^{w}\right)\} \cup \left\{c\left(v_{i}^{w} \mid v_{ij}^{o}\right) \mid 1 \leq j \leq m_{i}\right\}$ jointly restrict it, corresponding to all edges that points to $v_{i}^{w}$.
We formulate the coordinate determination as a \textit{nonlinear constrained optimization problem} (P1):
\begin{align}
\text{(P1): } & \underset{v_i^w}{\text{Maximize}} \sum_{c \in \mathcal{C}_{v_{i}^{w}}} \text{sum}(c) \\
& \text{subject to} \quad \text{min}(c) \geq 0, \ \forall c \in \mathcal{C}_{v_i^w} \\
& \qquad \qquad \quad \ \|v_i^w - x_j\| \geq L, \ \forall j \in \{1,...,k-1\}
\end{align}
where $ x_j $ represents the $ j $-th solution of $ v_{i}^{w} $.
The functions $\text{sum}(c)$ and $\text{min} (c)$ represent the sum and minimum values of the sub-constraints in $c$, respectively.
This constraint is solved for multiple times, and in the $k$-th iteration, we incorporate the first $ k-1 $ solutions into the constraint conditions.
For object nodes $v_{ij}^{o}$, we retain only the constraint conditions (without the optimization objective).
A pre-trained vision model perceives objects in the observation and the objects are projected onto the BEV map.
The coordinates of objects satisfying the constraint conditions are selected as solutions for $v_{ij}^{o}$.

\begin{table}[htbp]
    \centering
    \caption{Results of R2R-CE and RxR-CE in Habitat simulator. We mainly compare the SR and SPL of SOTA methods of VLN-CE.}
    \label{tab:main}
    \resizebox{\textwidth}{!}{
        \begin{tabular}{lcccccccccc}
            \toprule
            \multirow{3}{*}{\textbf{Method}} & \multirow{3}{*}{\textbf{Zero-shot}} & \multirow{3}{*}{\textbf{Training-free}} 
            & \multicolumn{4}{c}{\textbf{R2R Unseen}} & \multicolumn{4}{c}{\textbf{RxR Unseen}} \\
            \cmidrule(lr){4-7} \cmidrule(lr){8-11}
            & & & NE & OSR & \textbf{SR} & \textbf{SPL} & NE & OSR & \textbf{SR} & \textbf{SPL} \\
            \midrule
            Seq2Seq~\cite{krantz_vlnce_2020}   & $\times$ & $\times$ & 7.8 & 37.0 & 24.0 & 22.0 & - & - & -    & - \\
            WS-MGMap~\cite{chen2022weakly}  & $\times$ & $\times$ & 6.3 & 47.6 & 38.9 & 34.3 & 9.8 & 29.8 & 15.0 & 12.1 \\
            NaVid~\cite{zhang2024navid}     & $\times$ & $\times$ & 5.5 & 49.1 & 37.4 & 35.9 & 8.4 & 34.5 & 23.8 & 21.2 \\
            Uni-NaVid~\cite{zhang2024uninavid} & $\times$ & $\times$ & 5.6 & 53.3 & 47.0 & 42.7 & 6.2 & 55.5 & 48.7 & 40.9 \\
            ETPNav~\cite{an2024etpnav}    & $\times$ & $\times$ & 4.7 & 65.0 & 57.0 & 49.0 & 5.6 & - & 54.8 & 44.9 \\
            \midrule
            Cow~\cite{gadre2022cow}       & $\checkmark$ & $\times$ & - & - & 7.8  & 5.8 & - & - & 7.9  & 6.1 \\
            ZSON~\cite{majumdar2022zson}      & $\checkmark$ & $\times$ & - & - & 19.3 & 9.3  & - & - & 14.2 & 4.8 \\
            A²Nav~\cite{chen2023a2navactionawarezeroshotrobot}     & $\checkmark$ & $\times$ & - & - & 22.6 & 11.1 & - & - & 16.8 & 6.3 \\
            NavGPT-CE~\cite{zhou2023navgpt}     & $\checkmark$ & $\checkmark$ & 8.4 & 26.9 & 16.3 & 10.2 & - & - & -    & - \\
            CA-Nav~\cite{chen2024CANav}     & $\checkmark$ & $\checkmark$ & 7.6 & \textbf{48.0} & 25.3 & 10.8 & 10.4 & - & 19.0    & 6.0 \\
            InstructNav~\cite{InstructNav}     & $\checkmark$ & $\checkmark$ & \textbf{6.9} & - & 31.0 & \textbf{24.0} & - & - & -    & - \\
            \rowcolor{mygray}\textbf{GC-VLN (Ours)} & $\checkmark$ & $\checkmark$ & 7.3 & 41.8 & \textbf{33.6}    & 16.3    & \textbf{8.8} & \textbf{44.4} & \textbf{33.8} & \textbf{13.8} \\
            \bottomrule
        \end{tabular}
    }
\end{table}

\textbf{Navigation Tree.}
The constraint solver generates an uncertain number of solutions for nodes $v\in\mathcal{V}$.
We construct the navigation tree $\mathcal{T}$ to handle this uncertainty, where each node in $\mathcal{T}$ has a specific coordinate in the environment.
Assuming that the topological sort (TS) of $\mathcal{K}$ is $[v_{1}, v_{2}, ..., v_{|\mathcal{V}|}]$, the constraint solver progressively solves for the node coordinates.
According to the number of solutions from the constraint solver, TS is extended to a navigation tree, where the nodes at level $i$ in $\mathcal{T}$ are the coordinate solutions of the $i$-th node $v_{i}$ in $\mathcal{K}$.
Standing at the $i$-th level of $\mathcal{T}$, the number of branches in the $(i+1)$-th level of the navigation tree corresponds to the number of solutions for $v_{i+1}$.
The path from the root node in $\mathcal{T}$ to a leaf node in $\mathcal{T}$ corresponds to a specific path in the environment.

If the constraint solver fails to find a feasible solution, there will be no branches at that level for $v_{i}$, indicating that this particular navigation branch is unsuccessful.
In this case, the robot will backtrack in $\mathcal{T}$ until it finds a branch point with unexplored siblings.
At this branch point, the robot selects the next unexplored sibling as the future path.
The backtracking mechanism enhances the fault tolerance of GC-VLN, allowing the robot to explore potentially missed correct paths.
Navigation terminates if the robot reaches the last level of $\mathcal{T}$, corresponding to the endpoint $v_{|\mathcal{V}|}$.

\section{Experiments}

We conducted extensive experiments using simulators and real-world scenarios to validate the effectiveness of our method.
In this section, we will introduce our experimental setup, comparison with state-of-the-art methods, ablation studies, and qualitative analysis, respectively.

\subsection{Benchmarks and Implementation Details}

\noindent\textbf{Datasets:}
We conduct simulator experiments on the mainstream R2R-CE~\cite{anderson2018vision} and RxR-CE~\cite{ku2020room} datasets of VLN-CE tasks.
R2R-CE and RxR-CE are derived by converting the discrete trajectories from the R2R and RxR VLN datasets into continuous trajectories within the Habitat simulator~\cite{savva2019habitat}.
The scenes in the R2R-CE and RxR-CE datasets are sourced from the MP3D~\cite{Matterport3D} dataset.
We use the validation-unseen split of R2R-CE and RxR-CE, including 1,839 and 11,006 episodes.
The instructions in R2R-CE are entirely in English, while those in RxR-CE are in three languages.
The average path length and the instruction length in RxR-CE are greater than those in R2R-CE.

\noindent\textbf{Evaluation Metrics:}
Following~\cite{anderson2018evaluation,ilharco2019general}, we use \textit{success rate (SR)} and \textit{success rate weighted by path length (SPL)} as the main evaluation metrics.
SR represents the proportion of episodes where the agent successfully reaches within $m$ meters of the endpoint, where $m=3$.
SPL builds upon the success rate by incorporating path length, reflecting the similarity between the actual path and the ground truth path.
Besides, \textit{navigation error (NE)} and \textit{Oracle Success Rate (OSR)} are also reported.

\begin{table}
    \centering
    \caption{Effect of pipeline design in GC-VLN on R2R-CE benchmark.}
    \label{tab:ablation}
    \resizebox{\textwidth}{!}{
        \begin{tabular}{lcccc|lcccc}
            \toprule
            \multicolumn{5}{c|}{\textbf{Graph Constraint}} & \multicolumn{5}{c}{\textbf{Consrtraint Solver and Navigation Tree}} \\
            \textbf{Method} & NE & OSR & \textbf{SR} & \textbf{SPL} & 
            \textbf{Method} & NE & OSR & \textbf{SR} & \textbf{SPL} \\
            \midrule
            Relax constraints in $\mathcal{K}$   & 8.7 & 25.7 & 21.5 & 10.8 & Rearrange topological sort   & 9.0 & 36.4 & 26.2 & 12.1 \\
            Remove waypoint constraints          & 8.7 & 30.7 & 23.8 & 10.9 & Remove solution order   & 8.8 & 41.8 & 33.6 & 14.9 \\
            Remove object constraints   & 8.6 & 35.7 & 30.3 & 15.9 & Random Constraint Solver   & 9.1 & 9.1 & 7.0 & 2.6 \\
            Remove unary-constraint             & 8.7 & 37.9 & 32.6 & 16.3 & Simplify $\mathcal{T}$   & 8.9 & 38.2 & 29.6 & 11.4 \\
            Remove multi-constraint      & 8.7 & 36.9 & 30.9 & 15.2 & Remove Backtracking   & 9.0 & 23.2 & 21.3 & 13.6 \\
            \textbf{Full Approach} & \textbf{7.3} & \textbf{41.8} & \textbf{33.6}    & \textbf{16.3} & 
            \textbf{Full Approach} & \textbf{7.3} & \textbf{41.8} & \textbf{33.6}    & \textbf{16.3} \\
            \bottomrule
        \end{tabular}
    }
\end{table}

\noindent\textbf{Compared Methods:}
We compare with state-of-the-art methods for training-free VLN-CE.
NavGPT-CE~\cite{zhou2023navgpt,zhou2024navgpt2} is the NavGPT adapted version for continuous environments.
CA-Nav~\cite{chen2024CANav} is a constraint-aware zero-shot VLN-CE method.
InstructNav~\cite{InstructNav} is a generic instruction navigation method that is applicable to VLN-CE, ojbect-goal navigation, and demand-driven navigation.

\noindent\textbf{Implementation Details:}
We evaluate GC-VLN in the habitat simulator and real-world robot Hexmove.
We deploy DeepSeek-R1~\cite{deepseekai2025deepseekr1incentivizingreasoningcapability} as LLM for instruction decomposition and Grounded-SAM-2~\cite{ren2024grounded, ravi2024sam2segmentimages, ren2024grounding} for object perception. The local policy is the Fast Marching Method~\cite{sethian1996fast}.

\subsection{Comparison with State-of-the-art}

We compare GC-VLN with the state-of-the-art VLN-CE methods in three settings for VLN-CE: supervised, zero-shot and training-free in Table~\ref{tab:main}.
GC-VLN surpasses previous zero-shot methods, and surpassed SOTA training-free method InstructNav by 2\% success rate on the R2R-CE benchmark.
On the RxR-CE, we outperform all zero-shot methods for which performance has been reported.
In the supervised setting, we also outperform some methods, such as NaVid on RxR-CE.

In particular, RxR is more challenging than R2R.
However, our method maintains high performance on RxR, which demonstrates its stronger ability to handle complex instructions.

\begin{figure*}
  \centering
    \includegraphics[width=\linewidth]{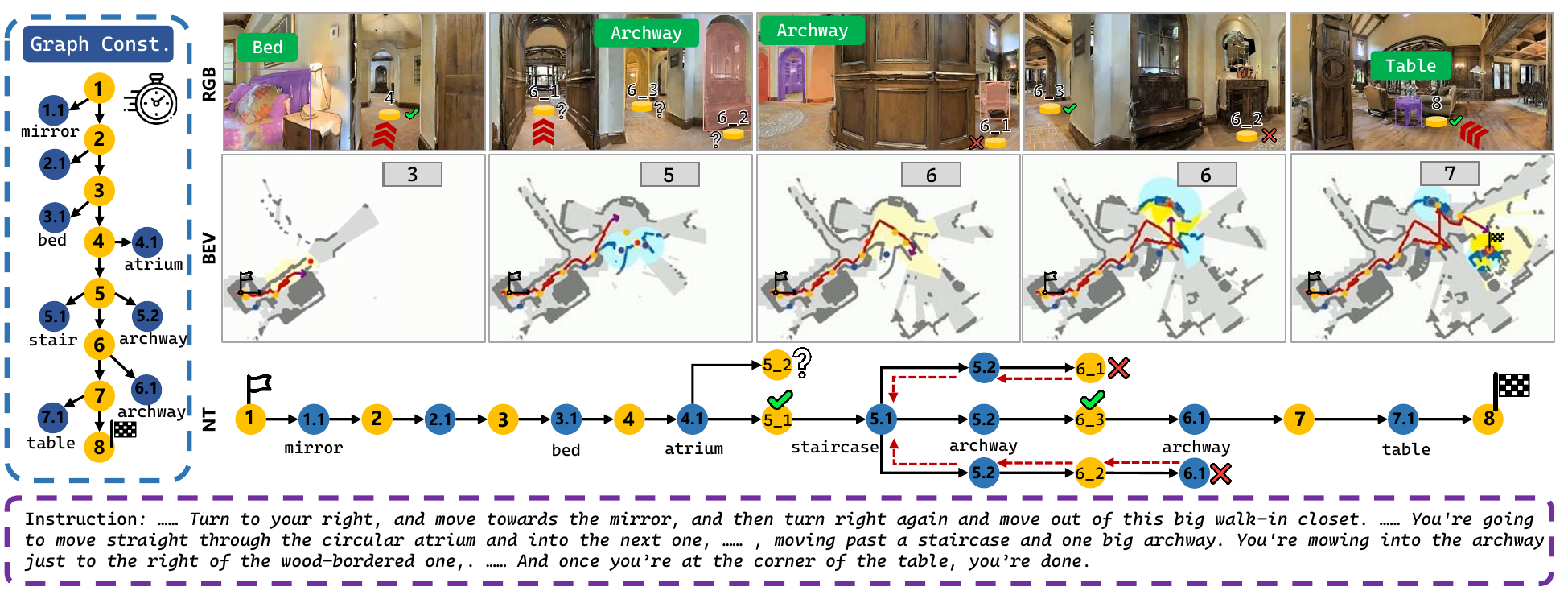}
  \caption{Demonstration of the graph constraints solving of GC-VLN. Here $t_i$ is the $i$-th branch node of the $t$-th level of navigation tree.}
  \label{fig:vis_sim}
\end{figure*}

\subsection{Ablation Study}

We conduct ablation studies on R2R-CE to validate the effectiveness of each module in GC-VLN.

\noindent\textbf{Effect of graph constraint:}
In Table~\ref{tab:ablation}, we first relax the graph constraints, which means removing the angle constraint and retaining only the maximum distance for the distance constraint.
The results show a significant drop in SR and SPL.
Then we remove the waypoint constraints $c \left( v_{t+1}^{w} \mid v_{t}^{w} \right)$, object constraints $c \left( v_{t+1}^{w} \mid v_{tj}^{o} \right)$, unary-constraints (type $1,2,4,5$) and multi-constraints (type $3,6$), respectively, replacing them with the weakest constraint (type $4$).
The performance of GC-VLN shows varying degrees of decline, demonstrating the effectiveness of graph constraints.

\noindent\textbf{Effect of constraint solver and navigation tree:}
We rearrange the topological sort by making object nodes no longer belong to a specific stage.
For solution order, we no longer use the order of coordinate solutions for a node to construct the navigation tree branches but adopt a random order.
For the constraint solver, instead of using the maximization objective function to solve coordinates, we randomly sample points within the region defined by the constraint conditions.
We simplify the navigation tree $\mathcal{T}$ by removing the earlier unexplored branches, and remove backtracking by not saving unexplored branches at all.
The performance declines across the board, demonstrating the effectiveness of constraint solver and navigation tree.

\begin{figure*}
  \centering
    \includegraphics[width=\linewidth]{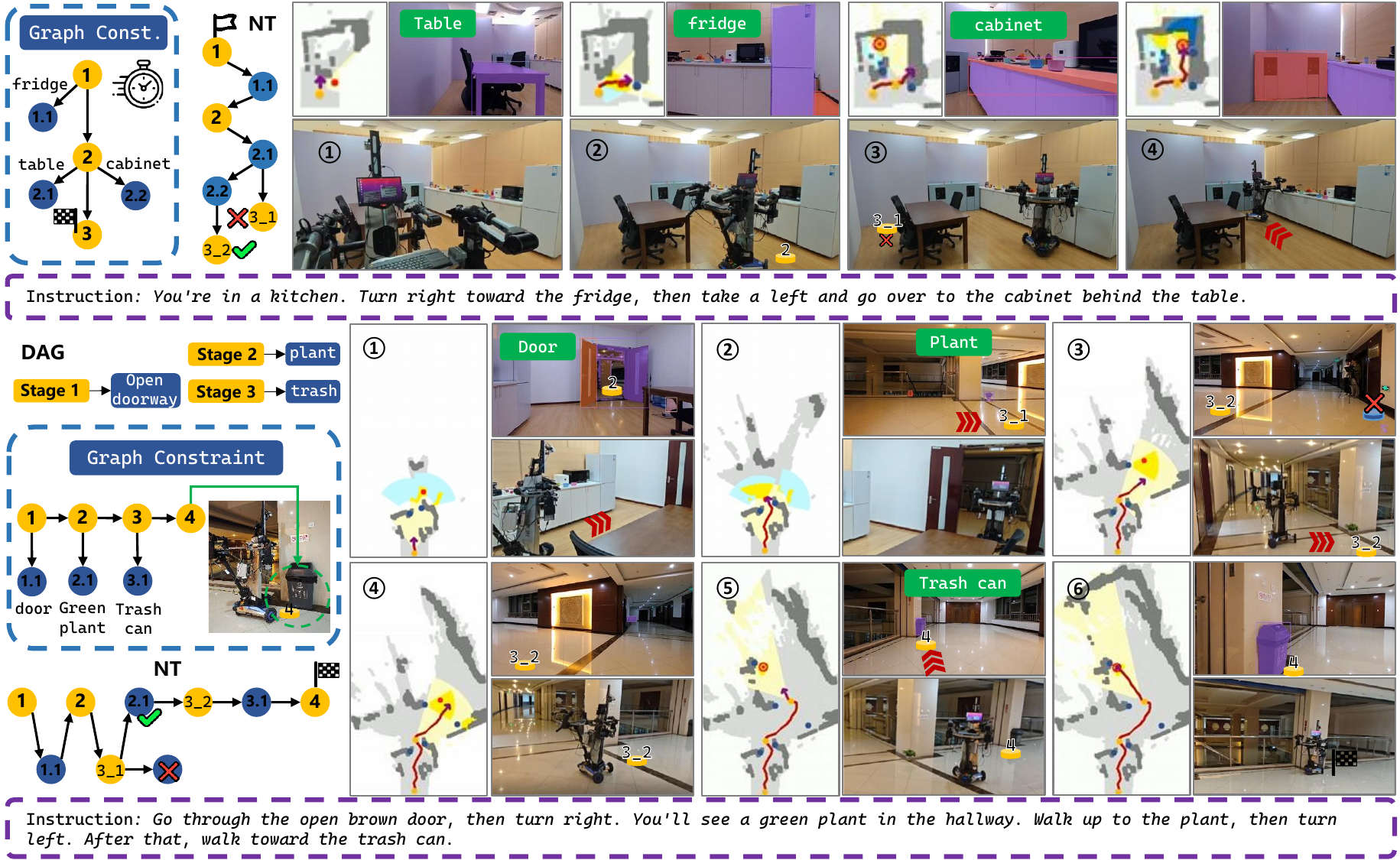}
  \caption{Demonstration of deployment in real-world environment.}
  \label{fig:vis_real}
\end{figure*}

\subsection{Qualitative Results}

To provide a more intuitive view of our method, we present visualizations of the navigation process in both the simulator and real-world environments.
As shown in Figure~\ref{fig:vis_sim}, the robot progressively explore the scene in the simulator and solve the position of each waypoint based on spatial constraints.
As illustrated in Figure~\ref{fig:vis_real}, we deploy GC-VLN in the real world, demonstrating its strong generalization ability in real-world deployment.
\section{Conclusion}

In this paper, we propose a graph constraint-guided training-free vision-and-language navigation framework.
Since zero-shot VLN methods for discrete environment and self-supervised VLN methods for continuous environment have a large sim-to-real gap, it is challenging to deploy them in real-world scenarios.
We construct a spatial constraint library that encompasses all possible spatial constraints and utilize LLM to decompose the instruction into a directed acyclic graph.
The graph is used to query the constraint library and obtain the graph constraints.
The positions of nodes in the directed acyclic graph are determined by the constraint solver, which solves the graph constraint optimization problem.
Navigation tree is used to handle the issue of uncertainty in the number of solutions.
Extensive experiments conducted in both simulated and real-world scenarios demonstrate the performance and generalization ability of GC-VLN, for which it can be easily deployed in real-world environment without performance degradation.

\clearpage
\acknowledgments{This work was supported in part by the Beijing Natural Science Foundation under Grant No. L247009, the National Natural Science Foundation of China under Grant 62125603, and the Beijing National Research Center for Information Science and Technology.}


\bibliography{main}  

\appendix
\clearpage
\setcounter{page}{1}

\section{Overview}
\label{sec:overview}
This supplementary material is organized as follows:
\begin{itemize}
\item Section \ref{sec:alg} provides the algorithm for the overall pipeline of GC-VLN.
\item Section \ref{sec:detail} provides the details of the approach.
\item Section \ref{sec:hard} provides the details of the hardware used in real-world experiments.
\item Section \ref{sec:exp} reports the results of additional ablation experiments.
\item Section \ref{sec:vis} shows visualization results of failure cases.
\item Section \ref{sec:prompt} details the prompts for LLM.
\end{itemize}

\section{Pipeline of GC-VLN}
\label{sec:alg}

In Algorithm~\ref{alg:block}, we provide an algorithm diagram of GC-VLN.

\begin{algorithm}[h]
\caption{Overall Pipeline of GC-VLN}
\begin{algorithmic}
\REQUIRE{Instruction $\mathcal{I}$, Observation $\mathcal{O}$}
\ENSURE{Goal Position $(x, y)$}
\STATE $\mathcal{G}\leftarrow \textrm{DecomposeInstruction}(\mathcal{I})$
\STATE $\mathcal{L}\leftarrow \textrm{ConstraintLibrary}()$
\STATE $\mathcal{C}\leftarrow \textrm{ConstructGraphConstraint}(\mathcal{G}, \mathcal{L})$
\STATE $\mathcal{T} \leftarrow$ NewNavigationTree()
\WHILE{True}
    \STATE $\{v_{i}\mid i=1,2,...,k\} \leftarrow$ ConstraintSolver($\mathcal{C},\mathcal{O}$)
    \STATE $\mathcal{T} \leftarrow$ UpdateNavigationTree($\mathcal{T},\{v_{i}\mid i=1,2,...,k\}$)
    \IF{$\{v_{i}\mid i=1,2,...,k\}==\Phi$}
        \STATE $v \leftarrow$ Backtrack($\mathcal{T}$)
    \ELSE
        \STATE $v \leftarrow$ GetWaypoint($\{v_{i}\mid i=1,2,...,k\}$)
    \ENDIF
    \STATE Go to $v\left(x,y\right)$
    \IF{$\textrm{isFinalWaypoint}\left(v\right)$}
        \STATE Stop at $v\left(x,y\right)$
    \ENDIF
\ENDWHILE
\end{algorithmic}
\label{alg:block}
\end{algorithm}

\section{Hardware Details}
\label{sec:hard}
\begin{wrapfigure}{r}{0.5\linewidth}
  \centering
    \includegraphics[width=\linewidth]{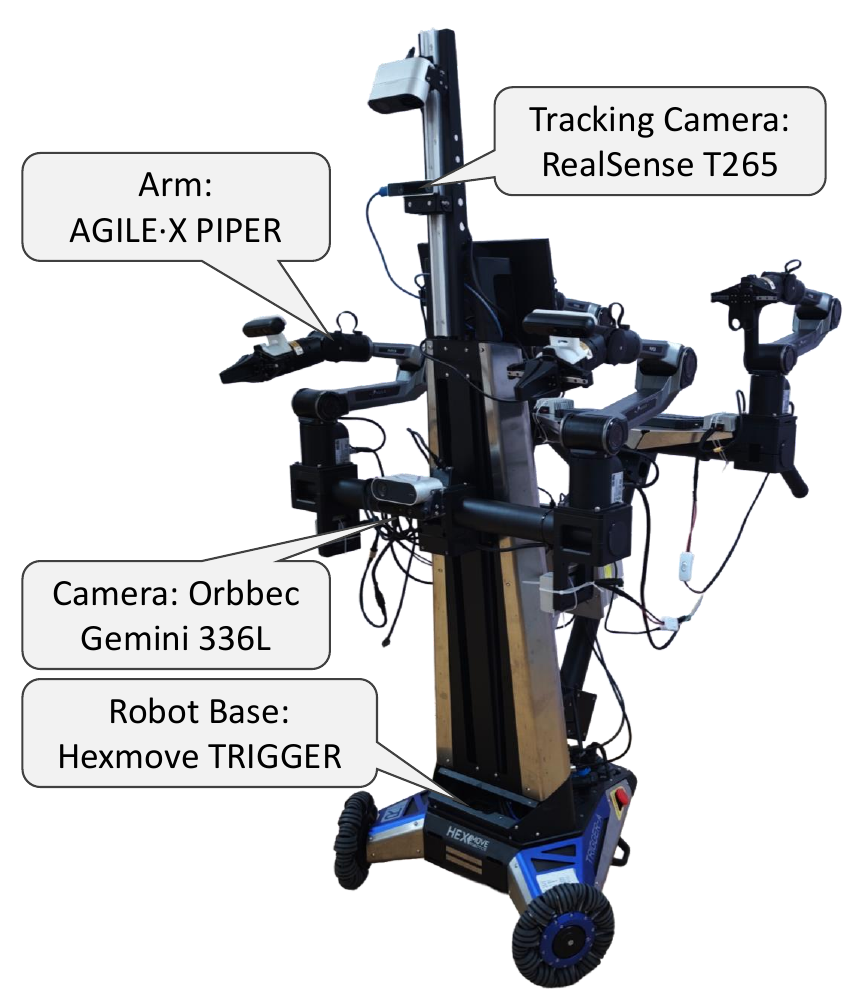}
  \caption{The robot employed for conducting the real-world experimental deployments.}
  \label{fig:robot}
\end{wrapfigure}

We present the hardware details of the robot employed in our real-world experiments in Figure~\ref{fig:robot}.
The robot base is the TRIGGER platform developed by Hexmove.
A monocular RGB-D camera serves as our input for observations, which is Orbbec Gemini 336L.
Our pose input is provided by the RealSense T265 tracking camera.
We utilize the PIPER robot arm from AGILE·X for object manipulation.

\section{Details of Approach}
\label{sec:detail}

\subsection{Instruction Decomposition}
LLM is prompted to decompose the linguistic instruction $\mathcal{I}$.
The decomposition of $\mathcal{I}$ must meet several rules:
1. Each stage must contain exactly ONE position change.
Rotating alone without movement is not a stage.
2. The direction of a stage is equal to the direction of the line from the start to the end of the stage.
3. Each stage comprises two attributes: position and a list of nodes.
Each node consists of two attributes: name and position.
4. "waypoint\_position" must belong to one of \textit{"front", "right", "left", "back", "unknown"}.
5. "object\_position" must belong to one of the types \textit{"right", "left", "through", "weave", "pass", "near", "back"}.

\subsection{Constraint Library}
As illustrated in Figure~\ref{fig:constraint} of the main text, based on the description in the instruction $\mathcal{I}$, the angle and distance information within a constraint can be identified, allowing the constraints to be classified into six types.
A unary constraint involves only two nodes, while a multi-constraint involves three or four nodes.
For example, the instruction "move forward 3 meters to the left" belongs to type 2, where both $u$ (blue node) and $v$ (green node) are waypoints.
The instruction "move forward through a door" corresponds to type 3, where $u_{1}$ (blue node) and $v$ (green node) are waypoints, while $u_{2}$ (gray node) is a door node.
The instruction "walk forward through the space between two chairs" falls under type 6, where $u_{1}$ (blue node) and $v$ (green node) are waypoints, and $u_{2}$ and $u_{3}$ (gray nodes) are chair nodes.

We formulate the sub-constraints for all six types of constraints.
Each type of constraint includes two possible sub-constraints: the angle constraint $c^{a}$ and distance constraint $c^{d}$.
The constraint types that include angular constraints are: 1, 2, 3, and 6.  
The constraint types that include distance constraints are: 2, 3, 5, and 6.
We formulate the constraint as:
\begin{align}
    c&=(\textbf{1}^{a}c^{a},\textbf{1}^{d}c^{d}), \quad \text{sum}(c)=\textbf{1}^{a}c^{a}+\textbf{1}^{d}c^{d}, \quad \text{min}(c)=\text{min}(\textbf{1}^{a}c^{a},\textbf{1}^{d}c^{d})
\end{align}
where $\textbf{1}^{a}$ and $\textbf{1}^{d}$ denote indicator functions representing the presence of angle and distance sub-constraints, respectively.

For types 1, 2, 4, and 5, the $c^{a}$ and $c^{d}$ are:
\begin{align}
    c^{a}\left(v \mid u\right)&=\cos (\Delta \phi)\|v-u\|-[\|v-u\|-(v-u) \cdot(\cos \phi, \sin \phi)]\\
    c^{d}\left(v \mid u\right)&=\Delta d^{2}-(\|v-u\|-d)^{2}
\end{align}

For type 3, the $c^{a}$ and $c^{d}$ are:
\begin{align}
    c^{a}\left(v \mid u_{1},u_{2}\right)&=\cos (\Delta \phi)\|v-u_{2}\|-[\|v-u_{2}\|-(v-u_{2}) \cdot(\cos \phi, \sin \phi)]\\
    c^{d}\left(v \mid u_{1},u_{2}\right)&=\Delta d^{2}-(\|v-u_{2}\|-d)^{2}
\end{align}
In type 1, 2, 3, 4, and 5, if the angle and distance are not explicitly specified, then $\Delta\phi=45^\circ$, $d=1.5m$ 
 and $\Delta d=1.5m$.

For type 6, the $c^{a}$ and $c^{d}$ are:
\begin{align}
    c^{a}\left(v \mid u_{1},u_{2},u_{3}\right)&=\cos (\Delta \phi)\|v-u_{1}\|-[\|v-u_{1}\|-(v-u_{1}) \cdot(\cos \phi, \sin \phi)]\\
    c^{d}\left(v \mid u_{1},u_{2},u_{3}\right)&=\Delta d^{2}-(\|v-u_{1}\|-d)^{2}\\
    \phi & =\arg \left(\frac{u_{2}-u_{1}}{\left|u_{2}-u_{1}\right|}+\frac{u_{3}-u_{1}}{\left|u_{3}-u_{1}\right|}\right) \\
    \Delta \phi & =\frac{1}{2} \arccos \left(\frac{\left(u_{2}-u_{1}\right) \cdot\left(u_{3}-u_{1}\right)}{\left|u_{2}-u_{1}\right| \cdot\left|u_{3}-u_{1}\right|}\right)
\end{align}

\subsection{Topological Sort}
To determine the order of nodes in graph constraint $\mathcal{K}$, we perform topological sort on the $\mathcal{K}$, which ensures that the parent node $u$ of each constraint $c \left(v \mid u \right)$ is always positioned ahead of its corresponding child node $v$.
The topological sort satisfies several conditions: 
1. The parent node must appear before its child node.
2. Among multiple child nodes of a single node, object nodes must precede waypoint nodes.
3. The order of multiple object child nodes under a single parent node must be consistent with the order in which they are mentioned in the instruction $\mathcal{I}$.

\section{Ablation Study}
\label{sec:exp}

In Table~\ref{tab:supp}, we report results of additional ablation experiments on hyperparameters of GC-VLN and RxR-CE benchmark.
For R2R-CE, we ablate the angle tolerance $\Delta\phi$ of the constraint type $1,2,3,6$, and the distance baseline $d$ of the all constraint types.
For RxR-CE, the settings of ablation are the same as those in the main text.

\begin{table}[h]
    \centering
    \caption{Effect of angle tolerance and distance baseline on R2R-CE. Effect of constraint and constraint solver on RxR-CE.}
    \label{tab:supp}
    \resizebox{\textwidth}{!}{
        \begin{tabular}{lcccc|lcccc}
            \toprule
            \multicolumn{5}{c|}{\textbf{Constraint Condition on Angle (R2R-CE)}} & \multicolumn{5}{c}{\textbf{Consrtraint Condition on Distance (R2R-CE)}} \\
            \textbf{Angle Tolerance $\Delta\phi$} & NE & OSR & \textbf{SR} & \textbf{SPL} & 
            \textbf{Distance Baseline $d$} & NE & OSR & \textbf{SR} & \textbf{SPL} \\
            \midrule
            $30^\circ$ & 10.2 & 32.6 & 30.1 & 15.9 & $0.8m$ & 10.6 & 35.7 & 28.5 & 14.7 \\
            \textbf{$45^\circ$ (Ours)} & \textbf{7.3} & \textbf{41.8} & \textbf{33.6}    & \textbf{16.3} & \textbf{$1.5m$ (Ours)} & \textbf{7.3} & \textbf{41.8} & \textbf{33.6}    & \textbf{16.3} \\
            $75^\circ$ & 10.2 & 33.8 & 31.5 & 15.6 & $2.5m$ & 10.1 & 39.0 & 30.5 & 14.0 \\
            \bottomrule
            \toprule
            \multicolumn{5}{c|}{\textbf{Graph Constraint (RxR-CE)}} & \multicolumn{5}{c}{\textbf{Consrtraint Solver (RxR-CE)}} \\
            \textbf{Method} & NE & OSR & \textbf{SR} & \textbf{SPL} & 
            \textbf{Method} & NE & OSR & \textbf{SR} & \textbf{SPL} \\
            \midrule
            Relax constraints in $\mathcal{K}$   & 10.8 & 31.5 & 24.5 & 9.9 & Random Constraint Solver & 11.1 & 21.9 & 18.0 & 10.6 \\
            \textbf{Full Approach} & \textbf{8.8} & \textbf{44.4} & \textbf{33.8} & \textbf{13.8} & \textbf{Full Approach} & \textbf{8.8} & \textbf{44.4} & \textbf{33.8} & \textbf{13.8} \\
            \bottomrule
        \end{tabular}
    }
\end{table}

\section{Visualization of Failure Cases}
\label{sec:vis}

\begin{figure}[h]
    \centering
    \includegraphics[width=\linewidth]{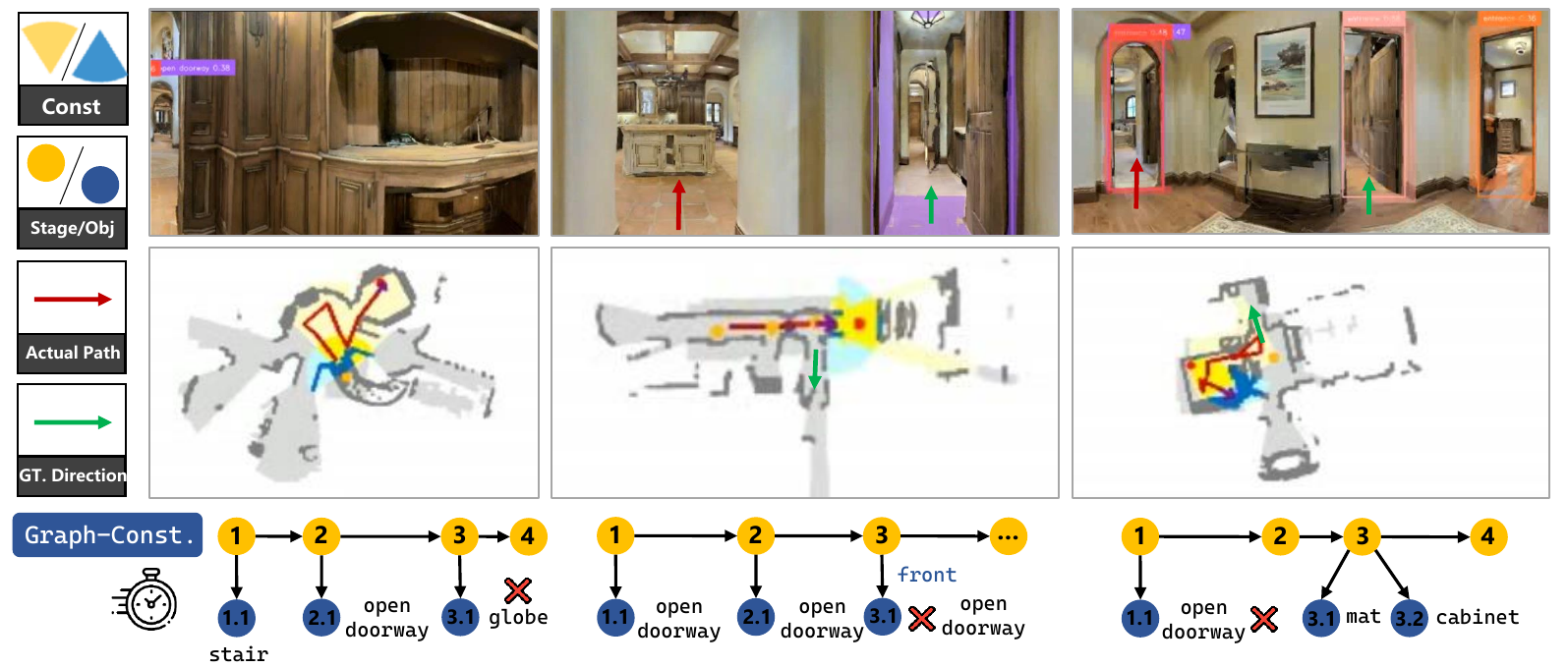}
    \caption{Visualization of three failure cases. In the first case, robot fails to locate the globe. In the second case, the robot mistakes "passing through the door to the right" for "forward" during the construction of the graph constraint. In the third case, the robot initially selects an incorrect path, and coincidentally encounters a correct object, which prevents timely backtracking.}
    \label{fig:vis_fail}
\end{figure}

In Figure~\ref{fig:vis_fail}, we further provide visualizations of failure cases for better understanding. 

\section{Prompts}
\label{sec:prompt}

We provide the prompt used for instruction decomposition in GC-VLN.

\begin{graybox}
    \texttt{Parse navigation instructions into movement stages in JSON format: \{}
    
    \texttt{\quad "stage 1": \{}

    \texttt{\quad \quad "waypoint position": <position>,}

    \texttt{\quad \quad "connected nodes": [}

    \texttt{\quad \quad \quad\{"node": <object>, "object position": <position>\},}
    
    \texttt{\quad \quad \quad ...}
    
    \texttt{\quad \quad ]}
    
    \texttt{\quad \},}

    \texttt{\quad ...}

    \texttt{\}}

    \texttt{Input Instruction: \textcolor{red}{<Instruction>}}
    
    \texttt{Rules:}
    
    \texttt{Each stage has one position change.
    The key "waypoint position" refers to the relative position of the next waypoint with respect to the current one, encompassing both the distance and the angle between them.
    The key "connected nodes" means the objects involved in the current stage.
    Each node includes two attributes: the "node" name and its "position" which denotes the relative position with respect to the waypoint.}

\end{graybox}

\vspace{-2mm}
where \texttt{\textcolor{red}{<Instruction>}} will be replaced by the input instruction.

\end{document}